\documentclass[pdflatex,sn-nature]{sn-jnl}% Style for submissions to Nature Portfolio journals
\usepackage{graphicx}%
\usepackage{multirow}%
\usepackage{amsmath,amssymb,amsfonts}%
\usepackage{amsthm}%
\usepackage{mathrsfs}%
\usepackage[title]{appendix}%
\usepackage{xcolor}%
\usepackage{textcomp}%
\usepackage{manyfoot}%
\usepackage{booktabs}%
\usepackage{cleveref}
\usepackage{algorithm}%
\usepackage{algorithmicx}%
\usepackage{algpseudocode}%
\usepackage{listings}%
\usepackage{booktabs}
\usepackage{pifont} 
\usepackage{tablefootnote}
\usepackage{makecell}
\lstdefinestyle{prompt}{
  basicstyle=\ttfamily\scriptsize,
  breaklines=true,
  columns=fullflexible,
  keepspaces=true,
  showstringspaces=false,
  frame=single
}

\newcommand{\xmark}{\ding{55}}

\theoremstyle{thmstyleone}%
\theoremstyle{thmstyletwo}%

\theoremstyle{thmstylethree}%

\begin{document}

\title[Article Title]{Advancing and Evaluating AI Agents for Frontier Physics Research}

% TODO(authors): designate the corresponding author(s) with \author* and add email
% addresses before submission.

%%=============================================================%%
%% GivenName	-> \fnm{Joergen W.}
%% Particle	-> \spfx{van der} -> surname prefix
%% FamilyName	-> \sur{Ploeg}
%% Suffix	-> \sfx{IV}
%% \author*[1,2]{\fnm{Joergen W.} \spfx{van der} \sur{Ploeg} 
%%  \sfx{IV}}\email{iauthor@gmail.com}
%%=============================================================%%

\author[1,2,8]{\fnm{Tingjia} \sur{Miao}}
\equalcont{These authors contributed equally to this work.}

\author[1]{\fnm{Wenkai} \sur{Jin}}
\equalcont{These authors contributed equally to this work.}

\author[1,3,4,5]{\fnm{Jinxin} \sur{Tan}}
\author[1,3,4,5]{\fnm{Muhua} \sur{Zhang}}
\author[1]{\fnm{Xianghe} \sur{Pang}}
\author[1]{\fnm{Zexi} \sur{Liu}}
\author[1]{\fnm{Yuwen} \sur{Du}}
\author[1]{\fnm{Tian} \sur{Jin}}
\author[3,4]{\fnm{Tu} \sur{Guo}}
\author[2,3,4]{\fnm{Zhengliang} \sur{Zhang}}
\author[2]{\fnm{Jingkun} \sur{Liu}}
\author[2,3]{\fnm{Yuelin} \sur{Hu}}
\author[2,3]{\fnm{Jiejun} \sur{Zhang}}
\author[1]{\fnm{Yunjie} \sur{Huang}}
\author[2,3]{\fnm{Yuhan} \sur{Wang}}
\author[3]{\fnm{Wenbo} \sur{Li}}
\author[2,3]{\fnm{Yinuo} \sur{Gao}}
\author[6]{\fnm{Shuo} \sur{Chen}}
\author[1]{\fnm{Rui} \sur{Ye}}
\author[7]{\fnm{Yuzhi} \sur{Zhang}}
\author[7]{\fnm{Linfeng} \sur{Zhang}}
\author*[6]{\fnm{Kun} \sur{Chen}} \email{chenkun@itp.ac.cn}
\author*[3,4,5]{\fnm{Wei} \sur{Wang}} \email{wei.wang@sjtu.edu.cn}
\author*[1]{\fnm{Weinan} \sur{E}} \email{weinan@math.pku.edu.cn}
\author*[1]{\fnm{Siheng} \sur{Chen}} \email{sihengc@sjtu.edu.cn}

\affil[1]{\orgdiv{School of Artificial Intelligence}, 
          \orgname{Shanghai Jiao Tong University}, 
          \orgaddress{\state{Shanghai}, 
                      \postcode{200030}, 
                      \country{China}}}

\affil[2]{\orgdiv{Zhiyuan College}, 
          \orgname{Shanghai Jiao Tong University}, 
          \orgaddress{\state{Shanghai}, 
                      \postcode{200240}, 
                      \country{China}}}

\affil[3]{\orgdiv{School of Physics and Astronomy}, 
          \orgname{Shanghai Jiao Tong University}, 
          \orgaddress{\state{Shanghai}, 
                      \postcode{200240}, 
                      \country{China}}}

\affil[4]{\orgdiv{Tsung-Dao Lee Institute}, 
          \orgname{Shanghai Jiao Tong University}, 
          \orgaddress{\state{Shanghai}, 
                      \postcode{200240}, 
                      \country{China}}}

\affil[5]{\orgdiv{State Key Laboratory of Dark Matter Physics}, 
          \orgname{Shanghai Jiao Tong University}, 
          \orgaddress{\state{Shanghai}, 
                      \postcode{200240}, 
                      \country{China}}}

\affil[6]{\orgdiv{Institute of Theoretical Physics}, 
          \orgname{Chinese Academy of Sciences}, 
          \orgaddress{\state{Beijing}, 
                      \postcode{100190}, 
                      \country{China}}}

\affil[7]{\orgname{DP Technology}, 
          \orgaddress{\state{Beijing}, 
                      \postcode{100082}, 
                      \country{China}}}

\affil[8]{\orgname{SciLand}, 
          \orgaddress{\state{Shanghai}, 
                      \postcode{200030}, 
                      \country{China}}}

%%==================================%%
%% Sample for unstructured abstract %%
%%==================================%%

 \abstract{Advances of LLMs in cognitive reasoning and tool-using capabilities reveal the potential of agentic science, where LLM agents conduct autonomous research and facilitate scientific discovery. Realizing this vision in frontier theoretical and computational physics, however, remains challenging because research workflows require domain expertise, long-horizon reasoning, and reliable numerical computation. To enable rigorous evaluation of agentic physics research, we construct \textsc{PRL-Bench}, a research-reproduction benchmark that faithfully captures the complexity of real-world physics research by distilling intricate research workflows into traceable artifacts and diverse evaluation rubrics. Adapted from 100 \emph{Physical Review Letters} papers spanning high-energy physics, condensed matter physics, AMO, and other frontiers. Each task is estimated by domain experts to require more than six hours for a specialized PhD student to reproduce independently. Evaluation on \textsc{PRL-Bench} reveals that existing agents systems remain inadequate for reliable long-horizon research workflows, motivating architectures specifically designed for reasoning and exploration in scientific research. We therefore present \textsc{PhysMaster}, a scientific agent integrating adaptive MCTS-based multi-trajectory exploration and hierarchical memory to improve robustness in extended research horizons and enable consistent knowledge accumulation. Comprehensive evaluation demonstrates that \textsc{PhysMaster} achieves the highest overall score of 51.08 on \textsc{PRL-Bench}, outperforming Codex, OpenHands, OpenClaw, and ReAct. Across diverse backbone models, \textsc{PhysMaster} consistently delivers substantial relative improvements ranging from 14.13\% to 93.38\%. Error analysis further shows that while existing agents frequently fail because of insufficient physics knowledge, unstable derivations, and the burden of long-horizon reasoning, \textsc{PhysMaster} substantially mitigates these limitations, particularly incomplete outcomes caused by extended research horizons. Together, \textsc{PRL-Bench} and \textsc{PhysMaster} establish a rigorous benchmark and an effective agentic system for advancing autonomous AI scientists in frontier physics research.}


\maketitle

\section{Introduction: Towards Agentic Scientific Research}
\label{sec:inrto}

\subsection{Benchmarking End-to-end Physics Research}

Advances of LLMs in cognitive reasoning, domain knowledge and tool-using capabilities reveal the potential of agentic science, where LLM agents conduct autonomous research rather than assist isolated tasks, thus facilitating scientific discovery. The paradigm of agentic science requires agents to conduct robust reasoning and engage in long-horizon, autonomous exploration. 

Theoretical and computational physics constitute a demanding yet ideal domain for agentic science and related benchmarking. Frontier researches in this area involve intensive long-horizon scientific reasoning and integrate theoretical analysis with numerical computation, while most of them can be conducted and verified entirely within digital environments, avoiding reliance on physical experiments. Agentic physics research tasks therefore provide a rigorous testbed for evaluating whether scientific agents can maintain reliable progress over extended research workflows. Such evaluations can further inspire advances aimed at stabilizing long-horizon exploration and enhancing theoretical grounding.

Current scientific benchmarks primarily evaluate knowledge comprehension and problem-solving ability, but rarely capture the exploratory and procedural nature of real scientific research. General expert-level benchmarks, including OlympiadBench, OlympicArena, OlymMath, and Humanity's Last Exam~\citep{he2024olympiadbench,huang2025olympicarena,sun2025olymmath,phan2025humanity}, offer rigorous assessments of advanced scientific and mathematical knowledge, while remain limited to comparatively constrained solution structures, and therefore only partially assess the capabilities required in extended scientific workflows.

Physics-oriented benchmarks provide closer alignment with real physics research by involving more specialized theoretical reasoning and settings related to physics research, including TPBench, PHYSICS, and PHYBench and RBench ~\citep{chung2025theoretical,feng2025physics,qiu2025phybench,qiu2026prbench}, as shown in table~\ref{tab:benchmark_comparison}. Nevertheless, existing physics benchmarks still mainly focus on problem solving or direct research reproduction, providing only limited coverage of the comprehensive, exploratory workflows characteristic of real scientific research.

\begin{table}[htbp]
\centering
\small
\begin{tabular}{lcccc}
\toprule
Benchmark & Scale & Knowledge Level & Research & Physics-Spec \\
\midrule
OlympicArena & Large (11163) & Undergraduate & \xmark & \xmark \\
HLE & Large (2500) & Graduate & \xmark & \xmark \\
PHYSICS & Large (1,297) & Undergraduate & \xmark & \checkmark \\
PHYBench & Large (500) & Graduate & \xmark & \checkmark \\
PRBench & Small (20) & Frontier Research & \checkmark & \checkmark \\
FrontierScience\tablefootnote{Denotes the research partition of the FrontierScience benchmark.} & Small (60) & Frontier Research & \checkmark & \xmark \\
\textsc{PRL-Bench} & Medium (100) & Frontier Research & \checkmark & \checkmark \\
\bottomrule
\end{tabular}
\caption{Comparison of representative benchmarks across scale, knowledge level, research orientation, and physics specialization.}
\label{tab:benchmark_comparison}
\end{table}

To address this gap, we present \textbf{PRL-Bench}, a high-fidelity research-reproduction benchmark that faithfully reflects the complexity of real-world physics research. It is adapted from 100 curated \emph{Physical Review Letters} papers and validated by domain experts, spanning six major subfield of modern physics: (i) AMO (ii) Astronomy \& Astrophysics, (iii) Condensed Matter Physics, (iv) High-Energy Physics, (v) Quantum Information, and (vi) Statistical Physics.

\textsc{PRL-Bench} is designed to preserve key features of real physics research, including nontrivial research motivations, theoretical derivations, numerical computation, open solution paths, long-horizon workflow structures, and diverse evaluation rubrics. Each task is estimated to require over 6 hours for a PhD student specializing in the corresponding research area to reproduce independently.

Evaluation on \textbf{PRL-Bench} demonstrates that existing agents remain insufficient for reliable autonomous physics research, especially when facing the complexity of extended scientific workflows. Therefore, developing scientific agents for reliable long-horizon exploration and continuous knowledge accumulation is essential. Such agents should be capable of maintaining coherent progress throughout extended workflows, incorporating external scientific knowledge when necessary, and progressively consolidating experience across research tasks.

\subsection{Towards Autonomous AI Physicist}

The advance of agentic science has motivated series of AI scientist systems, including Kosmos, Robin, and AI Scientist-v2, which combine literature grounding, tool use, iterative experimentation, and automated analysis to support scientific discovery~\citep{mitchener2025kosmos,ghareeb2025robin,yamada2025aiscientistv2}. In parallel, general-purpose agents have substantially improved autonomous task execution through tool use and search. ReAct-style agents interleave reasoning with external actions~\citep{yao2023react}, while systems such as Codex, OpenHands, and OpenClaw extend this paradigm with code execution, file manipulation, and iterative refinement~\citep{openai2025codex,wang2025openhands,openclaw2025}. Search-based agents further enhance long-horizon performance by exploring and evaluating multiple candidate trajectories~\citep{liu2025ml}. Despite these advances, existing systems remain primarily optimized for text-centric, well-defined, or comparatively short-horizon tasks, limiting their robustness on open-ended scientific research.

Domain-specific physics agents have begun to integrate agentic reasoning with specialized physics research workflows. GRACE performs simulation-guided search for particle-physics experiment design, while ColliderAgent automates collider-phenomenology workflows from theoretical specifications to simulated observables~\citep{hill2026grace,qiu2026collideragent}. However, these systems focus on specialized experimental or phenomenological pipelines, rather than the broader analytical and computational reasoning required across frontier theoretical and computational physics.

Toward agentic physics research, we build \textsc{PhysMaster}, a scientific research agent that integrates adaptive exploration and hierarchical memory to support reliable long-horizon research workflows. \textsc{PhysMaster} performs multi-trajectory exploration based on Monte Carlo Tree Search, enabling agents to evaluate alternative solution paths, dynamically revise intermediate failures, and fexibly reschedule during research workflows. Furthermore, \textsc{PhysMaster} introduces a hierarchical memory architecture that progressively consolidates execution experiences into compressed knowledge and transferable insights, allowing coherent reasoning across extended workflows while alleviating context limitations.

To further enhance flexibility and extensibility, \textsc{PhysMaster} incorporates \textsc{LANDAU}, a layered external knowledge system that integrates heterogeneous resources, including scientific literature, curated priors, reusable analytical skills, and expert workflows. This modular design enables the incorporation of evolving domain knowledge without modifying the underlying agent architecture. Together, adaptive exploration, hierarchical memory, and extensible knowledge integration provide a general framework for developing autonomous scientific research agents. Comprehensive evaluation on \textsc{PRL-Bench} demonstrates that \textsc{PhysMaster} consistently improves performance across all evaluated backbone models, achieving the highest overall score of 51.08 with relative gains ranging from 14.13\% to 93.38\%. Error analysis further indicates that the proposed architecture primarily improves the completeness and reliability of long-horizon research workflows, demonstrating its effectiveness and generalizability for agentic scientific research.

\subsection{Core Contributions}

The core contributions of this work are as follows:

\begin{enumerate}

\item \textbf{Providing comprehensive agentic physics research benchmark and rigorous evaluation}.  we construct \textbf{PRL-Bench}, a research-oriented benchmark with 100 agentic research tasks spanning six major subfield of modern physics, distilling intricate details of research into traceable artifacts and diverse rubrics. Rigorous quantitative evaluation have been conducted, demonstrating that lack of advanced physics knowledge, unstable derivations and burden of long-horizon are non-negligible for current research agents.

\item \textbf{Advancing the paradigm of autonomous research agents and agentic science}: We develop \textsc{PhysMaster}, a domain-specialized scientific research agent that introduces a generalizable architecture for autonomous research. Integrating hierarchical multi-agent collaboration with adaptive multi-trajectory exploration, \textsc{PhysMaster} enables reliable ultra-long-horizon exploration. Targeting physics research,\textsc{PhysMaster} incorporates specialized multi-layer knowledge base,and supports the integration of domain-specific scientific computing tools. Further, the proposed architecture is inherently transferable and provides a blueprint for building autonomous research agents across scientific disciplines, including computational chemistry, materials science, and engineering, etc.

\end{enumerate}

\section{Results}
\label{sec:results}

\subsection{Benchmark Overview}

Aiming at systematically and objectively assess the capability boundaries of scientific research agents, we construct \textsc{PRL-Bench}, a frontier research-reproduction benchmark that faithfully reflects the complexity of real-world physics research. It is adapted from 100 authoritative \textbf{Physical Review Letters} papers. Each paper is converted into several subtasks with traceable artifacts and diverse rubrics, thereby reconstructing the complete research workflow. Each task is estimated to require over 6 hours for a PhD student specializing in the corresponding research area to reproduce independently. A representative sample task is provided in Appendix~\ref{appendix:sample}.

In close collaboration with more than ten domain experts, we curated and adapted the source papers for benchmark construction. Selected papers emphasize theoretical derivation or numerical computation, while primarily experimental studies and tasks requiring inaccessible datasets, specialized simulation platforms, or prohibitive computational resources were excluded. All benchmark tasks were independently cross-validated by domain experts to ensure consistency with the scientific content and underlying physics of the source papers.

\textsc{PRL-Bench} distills intricate details of original research papers, preserving the nontrivial scientific motivation, nontrivial numerical computation and theoretical derivations, and long-horizon workflow structure of original papers. The benchmark tasks are deliberately adapted rather than directly adopting the exact results reported in the original papers, mitigating potential data leakage and enabling a more faithful and reliable evaluation of autonomous scientific research capabilities.

\textsc{PRL-Bench} also features \textbf{broad subfield coverage}, spanning six major subfields of modern physics: 

\begin{enumerate}
    \item Atomic Molecular and Optical Physics (AMO),
    \item Astronomy \& Astrophysics (Astro),
    \item Condensed Matter Physics (CMP),
    \item High Energy Physics (HEP),
    \item Quantum Information and Foundations (Quantum),
    \item Statistical Physics (Stat).
\end{enumerate}

\begin{figure}[htbp]
  \centering
  \includegraphics[width=0.5\linewidth]{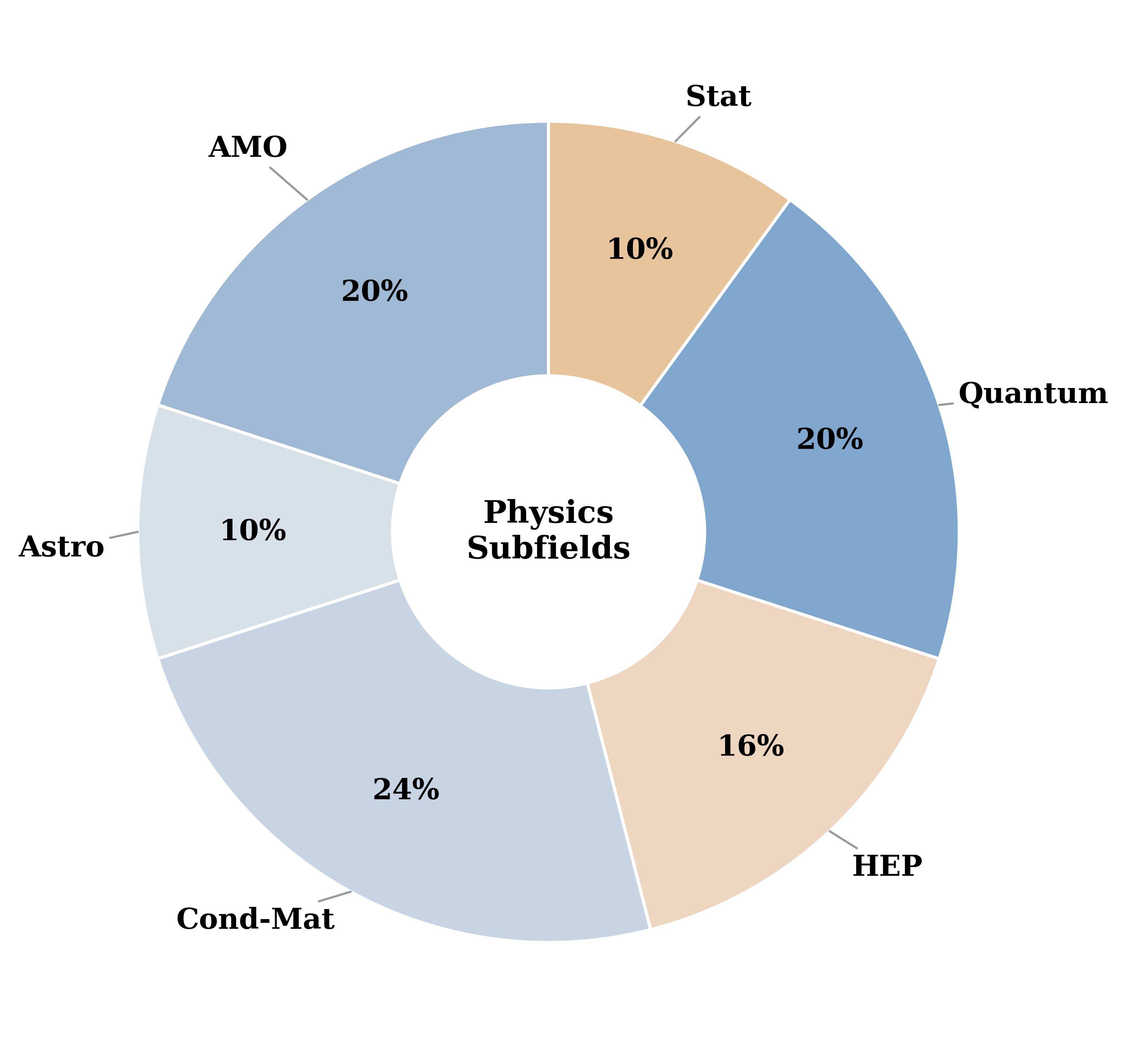}
  \caption{Subfield distribution of the 100 tasks in PRL-Bench.}
  \label{fig:benchmark-subfields}
\end{figure}

Collectively, these subfields cover physics from cosmological scales to the microscopic quantum regime and encompass diverse scientific methodologies, including theoretical derivation, phenomenological analysis, numerical computation, etc. The subfield distribution is shown in Fig.~\ref{fig:benchmark-subfields}. 

\subsection{Evaluation Protocol}

We evaluate a series of frontier LLMs as backbone models: GPT-5.4, Claude Opus 4.6, Gemini 3.1 Pro, Qwen-3.5-Plus, Kimi-K2.5, GLM-5.1, DeepSeek-V4-Pro, and DeepSeek-V4-Flash. 
All evaluations are conducted under identical task prompts, tool access, and scoring metric to ensure fair comparison.

\textbf{Agentic settings.} For each backbone model, we consider two agentic settings: (i) standard \textbf{ReAct} baseline following conventional reasoning--action--observation loop and (ii) our proposed \textsc{PhysMaster}. 

\textbf{Budgets.} To ensure fair comparison, the ReAct solver was capped at 50 reasoning iterations, and the core theoretician--critic exploration loop in \textsc{PhysMaster} was likewise limited to a maximum of 50 iterations.

\textbf{Tools.} All agents operate in an identical local execution environment with access to shell execution and file editing. To prevent data leakage from source papers or related solutions, all retrieval-based resources, including web search, external knowledge services (e.g., MCP) are disabled throughout evaluation.

\textbf{Scoring.} Each task is evaluated over three independent runs for each model, and the reported score is the average across runs to reduce stochastic variance. We use GPT-5.4 as an LLM judge to assess whether candidate solutions align with the rubrics. Scores from all subtasks are summed and normalized to a 0--100 scale.

\subsection{Consistent Gains of \textsc{PhysMaster} across Backbone Models}

We evaluate \textsc{PhysMaster} with eight backbone LLMs under identical task settings and compare it against their corresponding single-agent ReAct baselines. As shown in Table~\ref{tab:paired-results}, \textsc{PhysMaster} consistently improves performance across all 48 model--subfield combinations, demonstrating that the gains arise from the agentic architecture rather than outstanding backbone alone. 

\begin{table*}[htbp]
\centering
\caption{Performance comparison between ReAct baselines and corresponding \textsc{PhysMaster} counterparts on PRL-Bench. Scores are normalized to a 0--100 scale and averaged over three independent runs. Relative $\Delta$ denotes the improvement of \textsc{PhysMaster} over the corresponding ReAct baseline.}
\label{tab:paired-results}
\scriptsize
\setlength{\tabcolsep}{2.8pt}
\renewcommand{\arraystretch}{1.08}
\resizebox{\textwidth}{!}{
\begin{tabular}{llcccccccc}
\toprule
\textbf{Model} & \textbf{Agent} & \textbf{AMO} & \textbf{Astro} & \textbf{CMP} & \textbf{HEP} & \textbf{Quantum} & \textbf{Stat} & \textbf{Global} & \textbf{Relative $\Delta$} \\
\midrule
\multirow{2}{*}{GPT-5.4}
& ReAct & 40.46 & 39.05 & 42.99 & 39.48 & 56.10 & 49.06 & 44.76 & -- \\
& PhysMaster & 48.13 & 40.58 & 48.98 & 40.48 & 67.95 & 55.77 & \textbf{51.08} & +14.13\% \\
\addlinespace
\multirow{2}{*}{Claude-Opus-4.6}
& ReAct & 26.02 & 35.10 & 39.69 & 30.05 & 41.73 & 36.50 & 35.14 & -- \\
& PhysMaster & 35.97 & 46.52 & 46.70 & 38.01 & 64.03 & 59.13 & \textbf{47.94} & +36.43\% \\
\addlinespace
\multirow{2}{*}{Gemini-3.1-Pro}
& ReAct & 20.22 & 19.20 & 23.01 & 22.08 & 27.70 & 27.11 & 23.27 & -- \\
& PhysMaster & 31.82 & 24.46 & 32.98 & 27.12 & 46.82 & 30.98 & \textbf{33.52} & +44.05\% \\
\addlinespace
\multirow{2}{*}{Qwen-3.5-Plus}
& ReAct & 11.49 & 16.26 & 19.70 & 19.08 & 24.45 & 22.79 & 18.88 & -- \\
& PhysMaster & 27.50 & 32.40 & 36.82 & 30.28 & 49.44 & 41.85 & \textbf{36.50} & +93.38\% \\
\addlinespace
\multirow{2}{*}{Kimi-K2.5}
& ReAct & 19.70 & 18.43 & 21.99 & 22.98 & 30.37 & 25.88 & 23.40 & -- \\
& PhysMaster & 28.99 & 29.44 & 33.72 & 27.34 & 48.99 & 35.95 & \textbf{34.60} & +47.89\% \\
\addlinespace
\multirow{2}{*}{GLM-5.1}
& ReAct & 21.22 & 25.37 & 28.63 & 29.10 & 32.21 & 31.73 & 27.99 & -- \\
& PhysMaster & 36.18 & 37.87 & 41.05 & 34.54 & 61.39 & 47.28 & \textbf{43.40} & +55.06\% \\
\addlinespace
\multirow{2}{*}{DeepSeek-V4-Pro}
& ReAct & 29.37 & 32.83 & 37.03 & 33.84 & 41.83 & 48.19 & 36.65 & -- \\
& PhysMaster & 44.11 & 38.67 & 43.91 & 39.66 & 61.51 & 54.14 & \textbf{47.29} & +29.05\% \\
\addlinespace
\multirow{2}{*}{DeepSeek-V4-Flash}
& ReAct & 27.87 & 24.40 & 33.64 & 28.98 & 35.49 & 35.11 & 31.33 & -- \\
& PhysMaster & 36.29 & 36.29 & 39.80 & 36.15 & 56.98 & 50.84 & \textbf{42.70} & +36.31\% \\
\bottomrule
\end{tabular}
}
\end{table*}

At the global level, \textsc{PhysMaster} consistently outperforms ReAct across all backbone models, with relative improvements ranging from 14.13\% to 93.38\%. The strongest overall performance is achieved by GPT-5.4 with \textsc{PhysMaster} (51.08), followed by Claude Opus 4.6 (47.94) and DeepSeek-V4-Pro (47.29). The consistent gains across diverse backbone models indicate that the improvement mainly originates from the proposed agentic workflow rather than backbone capability alone. In particular, adaptive exploration and hierarchical memory provide additional support for maintaining reliable progress across long-horizon research processes.

Improvements also vary across physics subfields. For example, GPT-5.4 achieves gains ranging from 2.52\% in High-Energy Physics to 21.13\% in Quantum Information and Foundations, while remaining consistently positive across all six subfields. Such variation reflects differences in workflow characteristics across research domains. Tasks involving complex derivations and multiple intermediate reasoning stages benefit more from adaptive exploration and memory consolidation, whereas domains with relatively standardized solution procedures show smaller but still consistent improvements.

To conclude, these results indicate that \textsc{PhysMaster} provides a robust and generalizable framework for enhancing scientific research agents across diverse physics domains.

\subsection{Comparison with General Agent Baselines}

We compare \textsc{PhysMaster} with three representative general-purpose agent frameworks, Codex, OpenClaw, and OpenHands, using identical backbone (GPT-5.4) and evaluation settings. 

\begin{table*}[htbp]
\centering
\small
\caption{Comparison of GPT-5.4-based agent frameworks on PRL-Bench with network-connected tools disabled. Entries in the subfield and global rows are normalized rubric-based scores on a 0--100 scale, while the final row reports the average model-API cost per task in USD.}
\label{tab:prlbench-comparison}
\setlength{\tabcolsep}{5pt}
\renewcommand{\arraystretch}{1.12}
\resizebox{\textwidth}{!}{
\begin{tabular}{lcccc}
\toprule
\textbf{Subfield} & \textbf{OpenHands} & \textbf{OpenClaw} & \textbf{Codex} & \textbf{PhysMaster} \\
\midrule
Atomic, Molecular and Optical Ohysics & 32.68 & 41.25 & 43.92 & \textbf{48.13} \\
Astronomy \& Astrophysics & 34.23 & 36.74 & 40.08 & \textbf{40.58} \\
Condensed Matter Physics & 48.71 & 48.75 & \textbf{55.55} & 48.98 \\
High Energy Physics & 37.32 & \textbf{49.92} & 36.05 & 40.48 \\
Quantum Information and Foundations & 45.55 & 47.11 & 56.13 & \textbf{67.95} \\
Statistical Physics & 42.58 & 51.73 & 55.10 & \textbf{55.77} \\
\midrule
\textbf{Global} & 40.99 & 46.20 & 48.62 & \textbf{51.08} \\
\midrule
\textbf{Average cost/task (USD)} & 1.72 & 0.82 & 3.85 & 3.51 \\
\bottomrule
\end{tabular}
}
\end{table*}

As shown in Table~\ref{tab:prlbench-comparison}, \textsc{PhysMaster} achieves the highest overall score (51.08), outperforming Codex (48.62), OpenClaw (46.20), and OpenHands (40.99). It ranks first in four of the six physics subfields, including AMO, Astronomy \& Astrophysics, Quantum Information and Foundations, and Statistical Physics. Although general-purpose agents have demonstrated strong capabilities in reasoning, code execution, and tool interaction, they are primarily optimized for sequential task execution and remain limited when facing extended research workflows that require continuous exploration, intermediate verification, and accumulation of progress across multiple stages.

In contrast, \textsc{PhysMaster} introduces adaptive exploration and hierarchical memory mechanisms to support long-horizon scientific workflows. Multi-trajectory exploration enables the agent to maintain alternative solution paths, recover from unsuccessful directions, and reschedule toward more promising candidates. Meanwhile, hierarchical memory preserves intermediate research experiences and progressively consolidates useful information while compress the context. 

The cost analysis highlights the trade-off between inference cost and research capability. \textsc{PhysMaster} incurs higher API cost than OpenHands and OpenClaw due to its multi-agent collaboration and iterative exploration, but achieves substantially stronger performance. Compared with the strongest competing framework, Codex, \textsc{PhysMaster} attains a higher overall score (51.08 vs.\ 48.62) while reducing API cost by approximately 9\%. These results suggest that the proposed architecture improves the efficiency of inference computation, achieving better research performance without increasing cost relative to the strongest general-purpose baseline.

\subsection{Failures Analysis}

Further, we analyze how \textsc{PhysMaster} redistributes failures across tasks. Failures are categorized into three mutually exclusive types:

\begin{enumerate}
    \item \textbf{Coverage}: omitted answers, partial answers, or failure to cover task requirements, primarily reflecting insufficient adaptation to long-horizon task.
    \item \textbf{Analytical}: failures arising within the derivation chain, such as incorrect mathematical derivation and unsupported intermediate assumptions, primarily reflecting deficiencies in reasoning ability as well as hallucination issues.   
    \item \textbf{Conceptual}: fundamental interpretation of the scientific problem, inappropriate choice of theoretical models, formulas or assumptions, primarily reflecting insufficient domain knowledge in physics.
\end{enumerate}

\begin{figure}[htbp]
\centering
\includegraphics[width=1\linewidth]{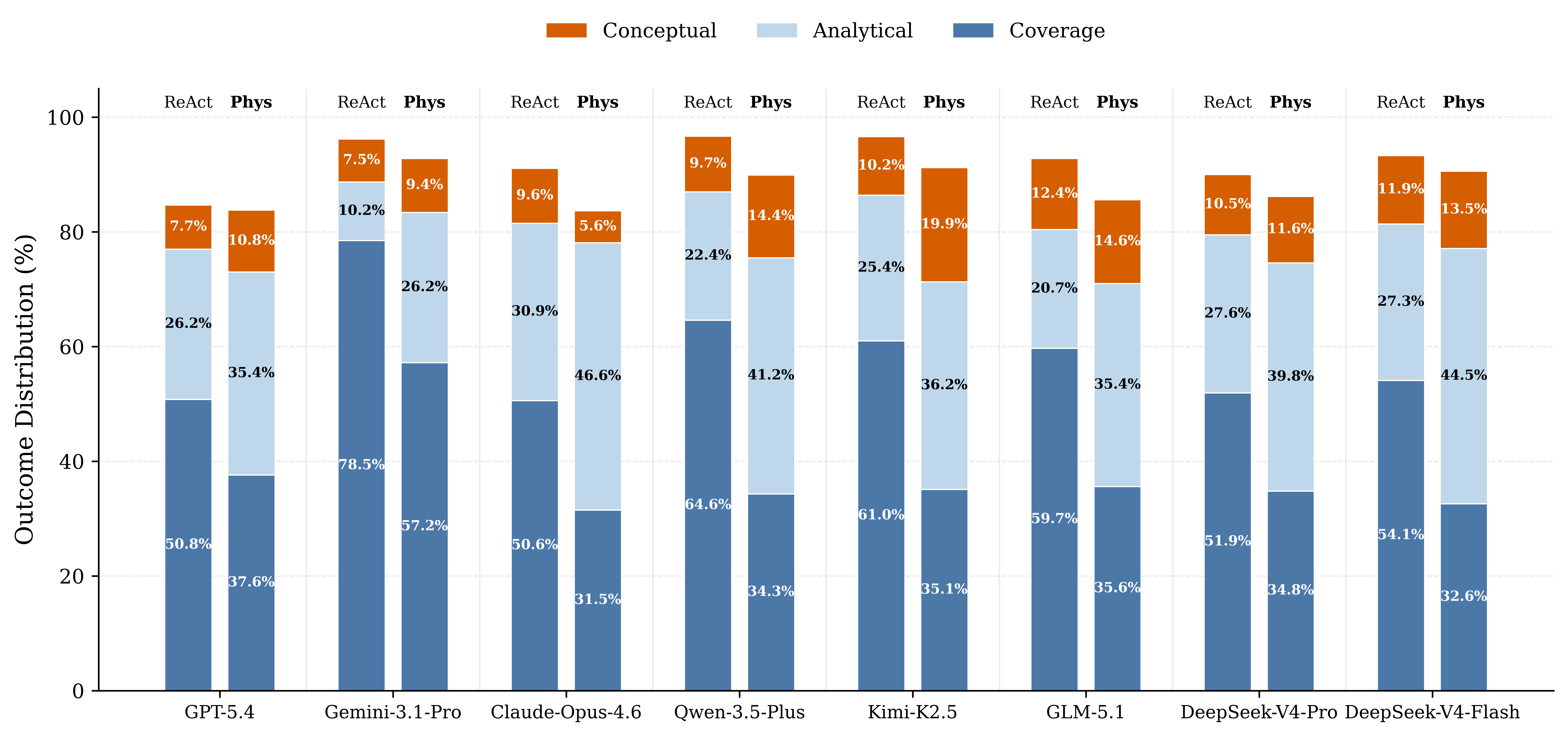}
\caption{Failure distributions of ReAct baselines and their \textsc{PhysMaster} counterparts.}
\label{fig:error-distribution}
\end{figure}

Figure~\ref{fig:error-distribution} reveals a clear redistribution of failure modes after applying \textsc{PhysMaster}. Across all eight backbone models, the most consistent change is a substantial reduction in Coverage errors. GPT-5.4, for example, reduces Coverage failures from 50.8\% to 37.6\%, and similar trends are observed for other models. This observation indicates that \textsc{PhysMaster} improves the completeness of long-horizon research workflows, enabling agents to execute more components of the intended scientific process instead of terminating after partial progress.

The reduction in Coverage errors is closely related to the proposed exploration and memory mechanisms. Adaptive exploration allows agents to maintain alternative research directions and recover from intermediate failures, reducing the risk of irreversible mistakes during long workflows. Meanwhile, hierarchical memory preserves intermediate findings and execution experiences, helping agents maintain consistency across multiple research stages. Together, these mechanisms mitigate the difficulties caused by extended research horizons.

As Coverage errors decrease, Analytical errors account for a larger proportion of the remaining failures across nearly all backbones. This redistribution reflects that agents are able to conduct more complete derivations, calculations, and code executions, exposing deeper reasoning and verification challenges that were previously masked by incomplete solutions. Meanwhile, Conceptual errors exhibit relatively modest changes, suggesting that while \textsc{PhysMaster} can mitigate the impact of limited domain knowledge through improved workflow organization, fundamental scientific understanding remains a persistent challenge.

Overall, the error distribution demonstrates that \textsc{PhysMaster} primarily improves autonomous research reliability by enabling more complete and coherent long-horizon workflows. The remaining bottlenecks are increasingly concentrated on deeper analytical reasoning and frontier-level scientific expertise, highlighting the need for further advances in both agent architectures and underlying foundation models.

\section{Discussion}
\label{sec:discussion}

This work presents a unified framework for advancing agentic physics research through both benchmarking and system design. PRL-Bench provides a research-oriented benchmark that preserves the open-ended reasoning, theoretical derivation, numerical computation, and long-horizon workflow characteristic of frontier physics research. Built from curated \emph{Physical Review Letters} papers and validated by domain experts, it offers a controlled testbed for systematically evaluating autonomous scientific research agents.

Across eight frontier backbone models, \textsc{PhysMaster} consistently improves performance over corresponding ReAct baselines and competitive general-purpose agent frameworks. These gains demonstrate that structured workflow orchestration can substantially enhance long-horizon scientific research without modifying the underlying language models. Error analysis further shows that the proposed architecture primarily reduces incomplete research workflows, enabling agents to execute a larger fraction of the intended scientific process while exposing analytical reasoning as the dominant remaining bottleneck.

At the same time, the benchmark also highlights important limitations of current research agents. Even the strongest systems continue to exhibit unstable long-horizon reasoning, imperfect theoretical derivations, and insufficient domain knowledge on frontier research problems. Moreover, \textsc{PRL-Bench} evaluates research-derived tasks with verifiable targets rather than open-ended scientific discovery, and therefore does not directly measure scientific novelty or creativity. Extending evaluation toward broader scientific domains, integrating stronger verification mechanisms and domain-specific computational tools, and improving reliable long-horizon exploration remain important directions for future work.

Overall, \textsc{PhysMaster} and \textsc{PRL-Bench} provide a practical paradigm of advancing agentic science and inspires long-horizon AI research agents towards more capable and reliable autonomous scientific discovery.

\section{Methods: \textsc{PhysMaster}, the Autonomous Physics Research Agent}
\label{sec3}

\subsection{Agentic Workflow and Hierarchical Agent Collaboration}

\textsc{PhysMaster} organize agentic physics research as a three-stage workflow with hierarchical agent collaboration. The core agents of \textsc{PhysMaster} and their responsibilities are demonstrated in Table~\ref{tab:components}.

\begin{table}[htbp]
  \caption{Core agents in \textsc{PhysMaster}.}
  \label{tab:components}
    \small
    \begin{tabular}{p{0.17\linewidth}p{0.70\linewidth}}
      \toprule
      Agent & Responsibility \\
      \midrule
      Clarifier & Reorganizes the original research query into an unambiguous task contract, preserving the core scientific motivation while making the essential requirements explicit. \\
      Supervisor & Directs search progress and allocates computation across alternative candidate solutions. \\
      Theoretician & Performs physical reasoning, numerical verification, and domain tool use to develop complete candidate solutions. \\
      Critic & Evaluates candidate soundness and task alignment, then guides subsequent exploration rounds. \\
      Summarizer & Summarize the candidate solution into the final report. \\
      \bottomrule
    \end{tabular}
  \vskip -0.08in
\end{table}

Prior to task execution, the Clarifier reorganizes the original research query into a structured task contract that preserves the core scientific motivation, resolves ambiguities, and makes the essential requirements explicit.

During the execution phase, the Supervisor, Theoretician, and Critic form an iterative exploration hierarchy. The Supervisor operates at the global control level, directing the search process and allocating computation across candidate solution paths; the Theoretician develops complete task-level candidates through physical reasoning, code-based verification, and tool use; and the Critic evaluates each candidate and guides subsequent search iterations.

Upon completion of the exploration phase, the Summarizer consolidates the selected candidate solution and its supporting artifacts into the final task deliverable. This decoupling of clarification, execution, evaluation, and integration externalizes workflow governance while preserving the coherence of the scientific output.

\begin{figure}[htbp]
  \centering
  \includegraphics[width=0.9\linewidth]{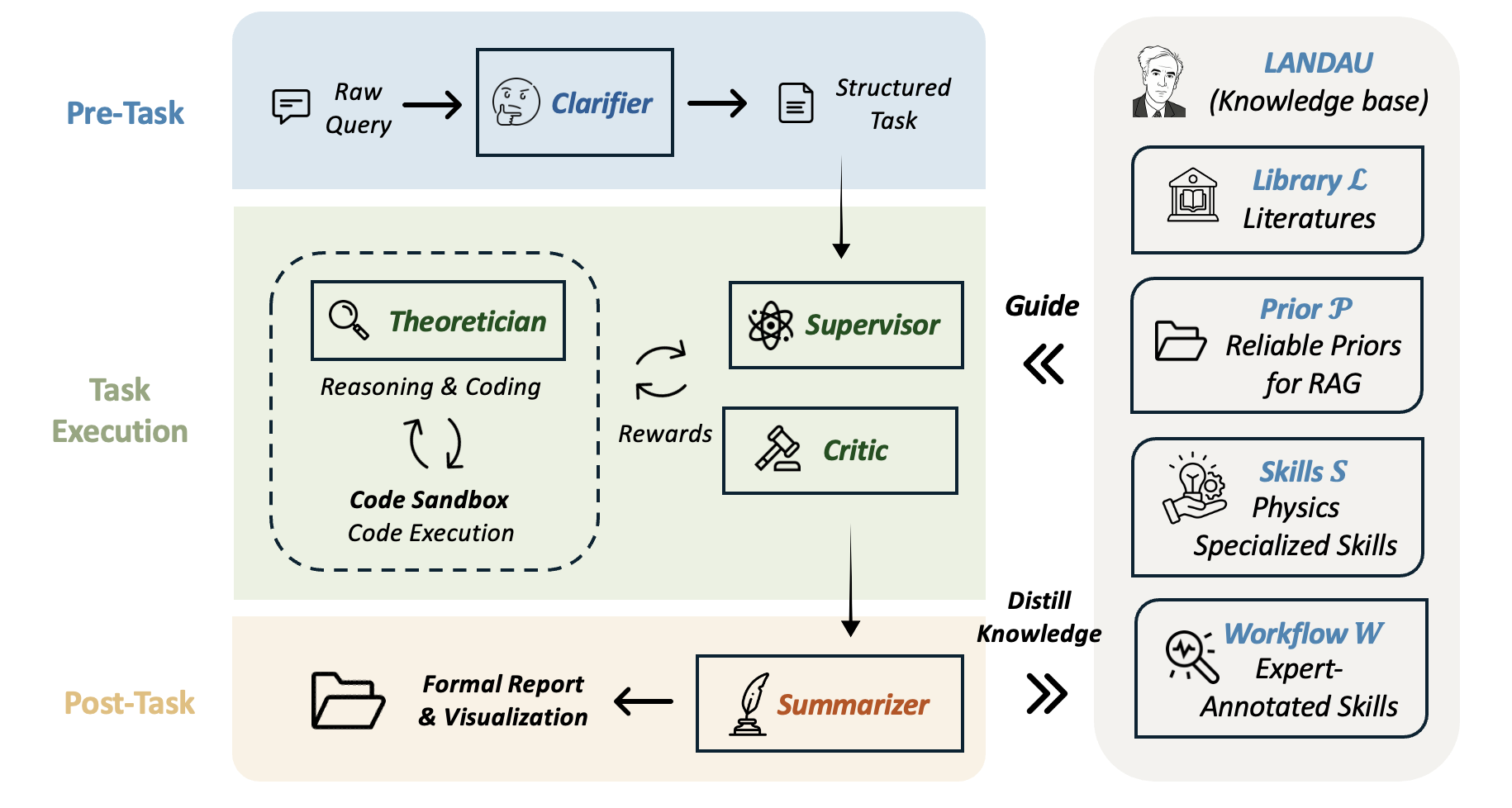}
  \caption{Hierarchical agent collaboration and core workflow of PhysMaster.}
  \label{fig:system-overview}
\end{figure}

\subsection{Adaptive Exploration Strategy and Hierarchical Memory System}

\subsubsection{MCTS-based Exploration}

\begin{figure}[htbp]
  \centering
  \includegraphics[width=0.8\linewidth]{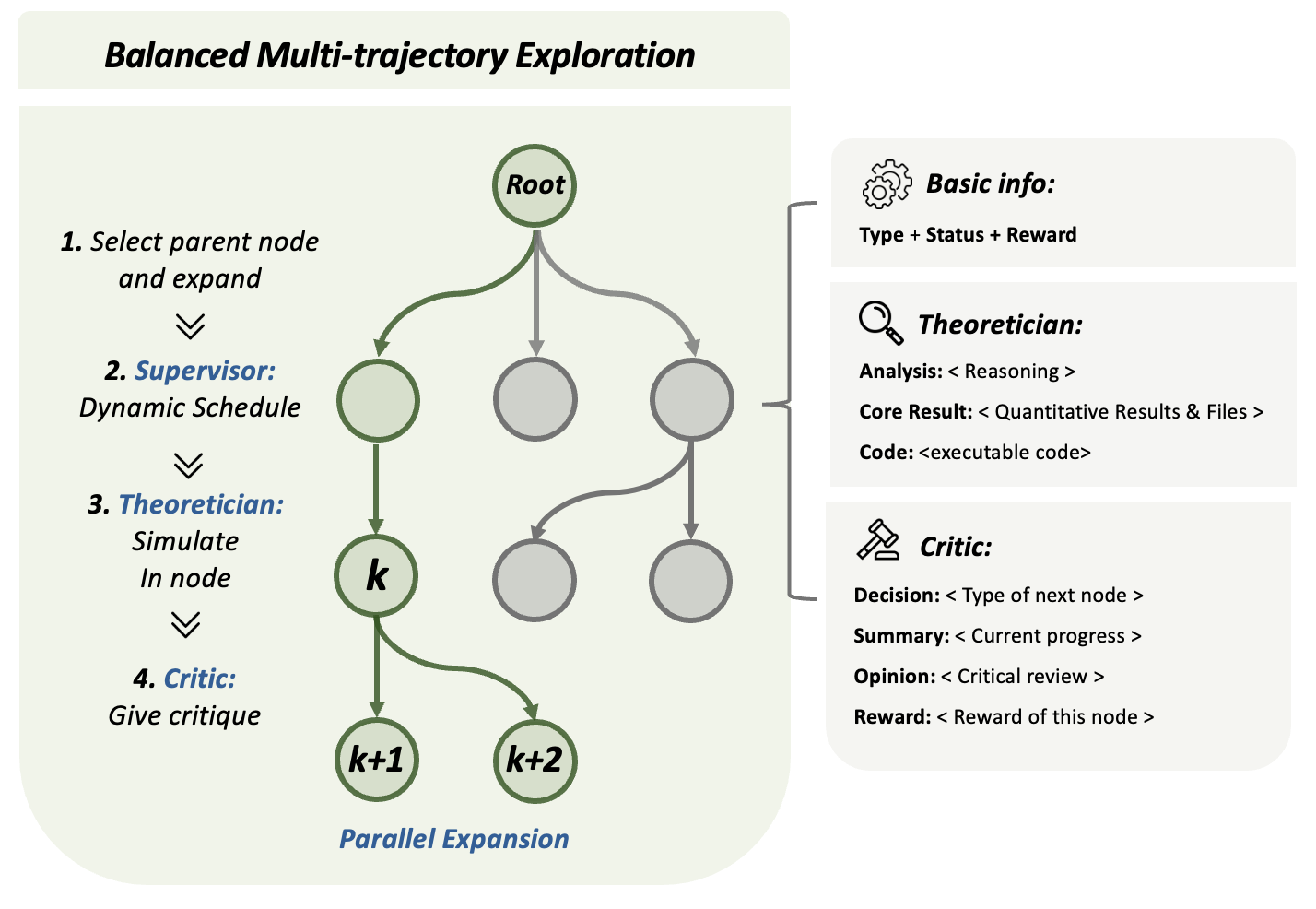}
  \caption{Critic-guided, MCTS-based multi-trajectory exploration in PhysMaster.}
  \label{fig:tree-search}
\end{figure}

Long-horizon physics tasks rarely admit a single obvious route from problem statement to final result. \textsc{PhysMaster} therefore uses Monte Carlo Tree Search (MCTS) as an external control structure for multi-trajectory exploration~\citep{kocsis2006bandit} (Fig.~\ref{fig:tree-search}). Each node represents a complete candidate result state rather than an isolated answer fragment. A \texttt{draft} expansion opens an alternative solution path, whereas a \texttt{revise} expansion improves an existing candidate in response to Critic feedback.

Node selection follows the upper confidence bound (UCB),
\begin{equation}
\mathrm{UCB}(v) = \frac{Q_v}{N_v}
+ C \sqrt{\frac{\ln(N_{\mathrm{parent}} + 1)}{N_v}},
\end{equation}
where $Q_v$ is the cumulative reward propagated through node $v$, $N_v$ is its visit count, $N_{\mathrm{parent}}$ is the visit count of its parent, and $C$ controls the exploration--exploitation trade-off. The first term favours branches with stronger historical returns, while the second preserves opportunities for less explored alternatives and reduces premature commitment to a locally plausible trajectory. Unvisited nodes are prioritized before the UCB score is applied.

Each search round consists of node selection, expansion, Critic evaluation, reward backpropagation, and optional pruning. The Supervisor selects a parent node using UCB and then determines whether to draft a new candidate or revise an existing one. The Critic assigns a reward and a search decision, and the reward is propagated to ancestor nodes to guide subsequent allocation of computation. Search terminates when a candidate is accepted as complete or when the configured round budget is exhausted.

\subsubsection{Hierarchical Cognitive Memory Across Tree Exploration}

To reconcile the finite context window with the large volume of information generated during long-horizon tree exploration, \textsc{PhysMaster} organizes memory into a hierarchical cognitive architecture: \emph{Per-Node Experience}, \emph{Compressed Knowledge}, and \emph{Cross-Task Wisdom}. Rather than preserving complete execution histories indefinitely, information is continuously consolidated as exploration proceeds. Detailed execution records are retained only while needed for local reasoning and verification, then distilled into reusable knowledge summaries, and finally promoted into transferable scientific wisdom after task completion.

\begin{enumerate}

\item \textbf{Per-Node Experience.}
Each search node maintains a complete record of its exploration process, including reasoning traces, tool interactions, generated code, execution logs, and intermediate computational results. This high-fidelity representation functions as working memory during active exploration, allowing failed derivations to be diagnosed, numerical results to be verified, and candidate solutions to be iteratively refined. Because these records preserve the full execution context, they provide the evidence required for subsequent evaluation and revision before any information is compressed.

\item \textbf{Compressed Knowledge.}
When a node reaches a sufficiently stable state, its detailed execution history is distilled into a concise knowledge summary attached to that node. Rather than storing complete reasoning traces, this representation retains key physical conclusions, successful and unsuccessful solution strategies, important implementation decisions, and lessons relevant to future revisions. These summaries are propagated through the search tree, allowing ancestor and sibling nodes to reuse accumulated insights without repeatedly processing verbose execution histories. This level therefore serves as a strategic memory shared across different branches of the ongoing search.

\item \textbf{Cross-Task Wisdom.}
After a task is completed, its consolidated knowledge is further abstracted into a persistent representation of transferable scientific experience. Instead of recording task-specific details, this level captures reusable reasoning patterns, methodological strategies, and domain knowledge that are applicable to related research problems. The accumulated wisdom is carried across tasks, enabling continual improvement over successive research episodes rather than treating each exploration as an isolated process.

\end{enumerate}

The hierarchy separates transient execution context from progressively more persistent cognitive state. As exploration advances, information is promoted from detailed execution records to reusable node-level knowledge and ultimately to transferable scientific wisdom according to its long-term reuse value. This progressive memory consolidation prevents context saturation while preserving the information required to support coherent long-horizon reasoning, enabling tree search to operate as a continual process of exploration, evaluation, consolidation, and revision rather than an ever-growing prompt history.

\subsection{LANDAU: Layered Academic Data Universe}

To support reliable research-time reasoning, \textsc{PhysMaster} is further equipped with \textsc{LANDAU}, short for \emph{Layered Academic Data Universe}, named in honor of the eminent physicist Lev Landau. LANDAU is a physics-specialized external knowledge system intended to provide stable support when parametric model memory alone is insufficient for frontier theoretical physics. Its composition is
\begin{equation}
\mathrm{LANDAU} = \mathcal{L} \cup \mathcal{P} \cup \mathcal{S} \cup \mathcal{W},
\end{equation}
where $\mathcal{L}$ denotes the library layer, $\mathcal{P}$ the prior layer, $\mathcal{S}$ the skill layer, and $\mathcal{W}$ the workflow layer.

\begin{enumerate}
    \item \textbf{Library layer $\mathcal{L}$.} This layer provides access to scientific papers and local literature collections. Its main role is to supply research evidence and frontier background that may not be stably covered by model pretraining alone.
    \item \textbf{Prior layer $\mathcal{P}$.} This layer stores high-confidence knowledge distilled from textbooks, lecture notes, and curated references. It serves as a stable source of conceptual constraints and theoretical frameworks. Typical authoritative textbooks represented in this layer are listed in Appendix~\ref{appendix:textbook}.
    \item \textbf{Skill layer $\mathcal{S}$.} This layer packages recurring analytical and computational routines into reusable execution units. Its role is to reduce repeated trial and error in familiar subtasks and make domain-specific operating practices available during task execution.
    \item \textbf{Workflow layer $\mathcal{W}$.} This layer provides high-level scaffolds for recurrent task classes. Its role is to help the system begin from a reasonable research structure and reduce early-stage drift in long-horizon tasks.
\end{enumerate}

In summary, LANDAU serves as a persistent and extensible knowledge base for agentic physics research. Its layered organization accommodates heterogeneous forms of domain expertise, including factual knowledge, methodological experience, research skills, and expert workflows, while allowing new knowledge sources to be incorporated without modifying the underlying reasoning architecture. This flexibility improves the reliability of long-horizon exploration and provides a general paradigm for equipping future research agents with diverse, evolving scientific knowledge.

\section{Data and Code Availability}

All source code for \textsc{PhysMaster} is available at \url{https://github.com/sjtu-sai-agents/PhysMaster}.

The full dataset of \textsc{PRL\_Bench} is available at \url{https://huggingface.co/datasets/AdrianMiao/PRL_Bench}.

% \section*{Acknowledgements}

%%=============================================%%
%% For presentation purpose, we have included  %%
%% \bigskip command. Please ignore this.       %%
%%=============================================%%
\bigskip

% APPENDIX
%%%%%%%%%%%%%%%%%%%%%%%%%%%%%%%%%%%%%%%%%%%%%%%%%%%%%%%%%%%%%%%%%%%%%%%%%%%%%%%
%%%%%%%%%%%%%%%%%%%%%%%%%%%%%%%%%%%%%%%%%%%%%%%%%%%%%%%%%%%%%%%%%%%%%%%%%%%%%%%
\newpage
\appendix
\onecolumn

\newpage

\section{Evaluation Prompt}

All candidate solutions are evaluated by an LLM judge using the prompt below with temperature set to zero. The parser extracts the numeric score normalized to 0--100, and parses the JSON object to take task-specific rubric results  and error annotations. Missing or unparseable scores are assigned zero. 

To ensure reproducibility, the complete evaluation prompts are shown below.

\begin{lstlisting}[style=prompt]
You are a strict evaluator for research-oriented physics tasks.

You will be given:
- Problem, including tasks/subtasks
- Rubrics
- Candidate answer

Evaluate the candidate strictly against the rubrics, which serve as the sole gold standard.

Scoring Rules:
- Assign a score normalized to a 0-100 scale.
- Each rubric item has equal weight unless explicit weights are provided.
- Scoring is strictly rubric-based: each rubric item is binary, satisfied or not satisfied.
- No partial credit, half-correct judgments, discretionary intermediate scoring, or approximate point allocation are allowed.
- Score the candidate by summing only the equally weighted rubric items that are correctly satisfied.
- Exact wording, identical derivations, or matching implementation details are not required.
- Derivations and formulations may use scientifically equivalent alternatives, but they must be logically valid and fully satisfy the rubric.
- Numerical results must strictly match the reference within the tolerance stated in the rubric/problem. If no tolerance is stated, use a default relative tolerance of 5%.
- For subjective or explanatory rubrics, apply a strict standard: the candidate must genuinely provide the required scientific content, not merely a vague or partially related discussion.
- For code-related rubrics, execution is not required, but the candidate code must be functionally equivalent to the rubric requirements and capable of fully satisfying the specified computational task. Only minor technical or implementation differences are allowed.

Error Types
- Conceptual: Fundamental misunderstandings of the underlying physics, including incorrect physical laws, theoretical models, core assumptions, or boundary/initial conditions. This category reflects errors in selecting or interpreting the appropriate physical framework ("using the wrong physics"), including unjustified assumptions, incorrect problem interpretation, or oversimplifying the task based on a flawed understanding.
- Analytical: Errors arising during derivation, reasoning, calculation, or implementation after the correct physical framework has been established. This includes logical flaws in derivations, algebraic mistakes, numerical or unit-conversion errors, and algorithmic or programming errors. The defining characteristic is that the physics is appropriate, but the execution of the solution is incorrect.
- Coverage: Required components explicitly requested by the problem statement that are entirely absent from the candidate answer, such as omitting a requested calculation, proof, justification, or one part of a multi-part question.

Distinguishing Coverage from Conceptual and Analytical
* If a required component is present but incorrect, classify it as Conceptual (incorrect physics, assumptions, or interpretation) or Analytical (errors in reasoning, derivation, calculation, or implementation).
* Use Coverage only when a required component is completely missing from the answer.
* Lack of rigor, incomplete justification, or shallow discussion should not be labeled as Coverage as long as the relevant topic is addressed. These deficiencies belong to Conceptual or Analytical depending on whether they reflect limitations in physical understanding or in the reasoning process. Coverage is strictly a binary assessment of presence versus absence.

Evaluation Requirements:
- Evaluate each task in the problem.
- For each task, strictly compare the candidate answer against that task's rubrics.
- List which rubric items are satisfied and which are not satisfied.
- If all rubric items for a task are satisfied, set error_type to "Correct".
- If any rubric item for a task is not satisfied, set error_type to the single dominant error type for that task.
- Provide a brief reason, 1-3 sentences, grounded only in the rubrics.
- The number of evaluated tasks must exactly match the tasks in the problem.

Output Format, followed exactly:

SCORE: <0-100, two decimal places>
{
  "tasks": [
    {
      "task_id": "...",
      "rubrics_satisfied": ["..."],
      "rubrics_not_satisfied": ["..."],
      "error_type": "...",
      "reason": "..."
    }
  ]
}

Constraints:
- error_type must be chosen strictly from the defined set.
- Do not include markdown, extra explanations, or fields beyond the specified output format.
\end{lstlisting}

\newpage
\section{Representative PRL-Bench Task}
\label{appendix:sample}

\subsection*{Introduction}

\textbf{Non-Abelian Chern Band in Rhombohedral Graphene Multilayers (Condensed Matter Physics, CMP)}

Moir\'e flat bands in rhombohedral multilayer graphene host interaction-driven
topological phases. At $\nu=2$, Hartree-Fock calculations reveal a doubly
spin-degenerate Chern band with $|C|=1$ and a non-commutative SU(2) gauge
structure---a non-Abelian Chern band, captured by the minimal model
$H = p^2/(2m) + 2V_0\sum_{i=1}^3 \sigma_i\cos(\mathbf{G}_i\!\cdot\!\mathbf{r})$.
A non-symmorphic symmetry $U_i = \sigma_i T_{\mathbf{L}_i/2}$ enforces exact
twofold band degeneracy and SU(2) Wilson loop holonomies; the occupied doublet
forms a skyrmionic spin texture with winding number $+2$.
Units: $\hbar^2/(2m)=1$, $L=1$, $V_0=0.05$.

% ============================================================
\subsection*{Task 1: Symmetry Analysis (Analytical)}

Consider the minimal Hamiltonian
$H = p^2/(2m) + 2V_0\sum_{i=1}^3\sigma_i\cos(\mathbf{G}_i\!\cdot\!\mathbf{r})$,
with $\mathbf{G}_i = G(\cos\theta_i,\sin\theta_i)$,
$\theta_i = 2\pi(i-1)/3 - \pi/6$, $G = 4\pi/(\sqrt{3}L)$, and lattice vectors
$\mathbf{L}_i = L(\cos(\theta_i-\pi/2),\sin(\theta_i-\pi/2))$ satisfying
$\mathbf{L}_i\!\cdot\!\mathbf{G}_j = 0$ ($i=j$), $\pm2\pi$ ($i\neq j$).

\begin{enumerate}
\item Define $U_i = \sigma_i T_{\mathbf{L}_i/2}$. Prove $[H,U_i]=0$,
  establish the algebra $\{U_i,U_j\}=2\delta_{ij}T_{(\mathbf{L}_i+\mathbf{L}_j)/2}$
  and $U_i^2=T_{\mathbf{L}_i}$, and conclude $U_i$ anticommute for $i\neq j$.

\item From eigenvalues $\lambda_{i,\pm}= \pm e^{i\mathbf{k}\cdot\mathbf{L}_i/2}$,
  prove exact twofold band degeneracy at every $\mathbf{k}$-point.

\item Show eigenvalue swapping along $\Gamma\to\mathbf{G}_j$ ($j\neq i$):
  $\lambda_{i,\pm}(0)=\pm1 \mapsto \lambda_{i,\pm}(\mathbf{G}_j)=\mp1$ via
  $e^{i\mathbf{G}_j\cdot\mathbf{L}_i/2}=e^{i\pi}=-1$, implying three distinct
  SU(2) fluxes ($\tau_1,\tau_2,\tau_3$) threading the BZ.

\item Trace the Hartree-Fock origin: Fock exchange
  $V_{\mathrm{Fock}}\sim\sum_i\sigma_i\langle S_i\rangle$ with spin-density-wave
  order at $\mathbf{G}_i$ yields $\sigma_i\cos(\mathbf{G}_i\!\cdot\!\mathbf{r})$;
  Hermiticity and threefold symmetry enforce equal $V_0$.
\end{enumerate}

\medskip
\noindent\textbf{Rubrics:}
\begin{enumerate}
\item $H = (p_x^2+p_y^2)/(2m)
  + 2V_0[\sigma_1\cos(\mathbf{G}_1\!\cdot\!\mathbf{r})
  + \sigma_2\cos(\mathbf{G}_2\!\cdot\!\mathbf{r})
  + \sigma_3\cos(\mathbf{G}_3\!\cdot\!\mathbf{r})]$ with
  $\mathbf{L}_i\!\cdot\!\mathbf{G}_j = 0$ ($i=j$), $\pm2\pi$ ($i\neq j$).
\item Commutation: $U_i\sigma_j\cos(\mathbf{G}_j\!\cdot\!\mathbf{r})U_i^{-1}
  = \sigma_i\sigma_j\sigma_i\cos(\mathbf{G}_j\!\cdot\!(\mathbf{r}+\mathbf{L}_i/2))$.
  $i=j$: $\mathbf{G}_i\!\cdot\!\mathbf{L}_i/2=2\pi$, sign $(+1)$.
  $i\neq j$: $\sigma_i\sigma_j\sigma_i=-\sigma_j$,
  $\mathbf{G}_j\!\cdot\!\mathbf{L}_i/2=\pm\pi$, sign $(-1)\!\times\!(-1)=(+1)$.
\item $\{U_i,U_j\}=2\delta_{ij}I_2T_{(\mathbf{L}_i+\mathbf{L}_j)/2}$,
  $U_i^2=T_{\mathbf{L}_i}$, eigenvalues $\pm e^{i\mathbf{k}\cdot\mathbf{L}_i/2}$.
\item $|\psi_{\mathbf{k}}\rangle$ and $U_i|\psi_{\mathbf{k}}\rangle$ orthogonal
  (opposite $U_i$ eigenvalues) $\Rightarrow$ exact twofold degeneracy.
\item $\mathbf{G}_j\!\cdot\!\mathbf{L}_i/2=\pm\pi$ ($j\neq i$)
  $\Rightarrow$ eigenvalue swap $\Rightarrow$ non-Abelian SU(2) holonomy.
\item Fock exchange with spin-off-diagonal $P(\mathbf{k})$ at $\mathbf{G}_i$
  generates $\sigma_i e^{i\mathbf{G}_i\cdot\mathbf{r}}$; equal $V_0$ from Hermiticity.
\end{enumerate}

% ============================================================
\subsection*{Task 2: Band Topology (Numerical)}

Diagonalize $H$ in the plane-wave basis
$|\mathbf{k}+n_1\mathbf{G}_1+n_2\mathbf{G}_2,\sigma\rangle$ with
$|n_i|\leq N_{\mathrm{cut}}=5$ (dimension 242).
Kinetic term: $H_{\mathrm{kin}} =
\hbar^2|\mathbf{k}+\mathbf{G}|^2/(2m)\,\delta_{\mathbf{G},\mathbf{G}'}
\delta_{\sigma,\sigma'}$.
Potential couples $\mathbf{G}\leftrightarrow\mathbf{G}\pm\mathbf{G}_i$ with
$V_0\sigma_i$, where $\mathbf{G}_1\!\leftrightarrow\!(1,0)$,
$\mathbf{G}_2\!\leftrightarrow\!(0,1)$, $\mathbf{G}_3\!\leftrightarrow\!(-1,-1)$.

\begin{enumerate}
\item Compute band structure along $\Gamma(0,0)\!\to\!K(\frac{2\mathbf{G}_1+\mathbf{G}_2}{3})\!
  \to\!M(\frac{\mathbf{G}_1}{2})\!\to\!\Gamma$ ($N_{\mathrm{seg}}=80$).
  Verify exact twofold degeneracy (max.\ splitting $<10^{-10}$).
  Report band gap at $\Gamma$ (lowest to next doublet) and lowest-doublet bandwidth.

\item Compute SU(2) Wilson loops $\Gamma\to\mathbf{G}_i$ ($i=1,2,3$,
  $N_{\mathrm{steps}}=200$) with parallel-transport gauge fixing:
  SVD of $M=\Psi_i^\dagger\Psi_{i+1}=U\Sigma V^\dagger$,
  rotate $\Psi_{i+1}\!\to\!\Psi_{i+1}VU^\dagger$,
  $W_i=\prod_j\Psi_j^\dagger\Psi_{j+1}^{\mathrm{fixed}}$.
  Verify: (i)~$\|W_i^2+I\|<0.05$, (ii)~$\det(W_i)\in[-1.01,-0.99]$,
  (iii)~$\|W_iW_j+W_jW_i\|<0.05$ for $i\neq j$.

\item At $\Gamma$, construct $U_i^{\mathrm{full}}
  = \sigma_i\otimes\operatorname{diag}(e^{i\mathbf{G}\cdot\mathbf{L}_i/2})$,
  project to doublet: $U_i^{\mathrm{proj}}
  = \Psi_\Gamma^\dagger U_i^{\mathrm{full}}\Psi_\Gamma$.
  Verify $\|\{U_i^{\mathrm{proj}},U_j^{\mathrm{proj}}\}\|<10^{-6}$ ($i\neq j$),
  eigenvalues $\{\pm1\}$ within $10^{-6}$.

\item Full BZ boundary loop
  $\Gamma\!\to\!\mathbf{G}_1\!\to\!\mathbf{G}_1\!+\!\mathbf{G}_2\!
  \to\!\mathbf{G}_2\!\to\!\Gamma$ ($N_{\mathrm{steps}}=100$/seg).
  Verify $W_{\mathrm{boundary}}\approx -I_{2\times2}$,
  eigenvalue phase sum $=2\pi$ ($C=1$).
\end{enumerate}

Produce one single-panel band structure figure ($\Gamma$-$K$-$M$-$\Gamma$,
lowest six doublets). Report band gap, bandwidth, Wilson loop matrices, and
symmetry operator properties as printed output.

\medskip
\noindent\textbf{Rubrics:}
\begin{enumerate}
\item $H$ dimension $2\times(2N_{\mathrm{cut}}+1)^2=242$; $H_{\mathrm{kin}}$
  diagonal, $H_{\mathrm{pot}}$ adds $V_0[\sigma_s]_{\sigma,\sigma'}$ for
  $\pm\mathbf{G}_s$ connections. $N_{\mathrm{seg}}=80$ along
  $\Gamma$-$K$-$M$-$\Gamma$.
\item Max.\ doublet splitting $<10^{-8}$ at all $\mathbf{k}$-path points.
\item $W_1$ ($\Gamma\!\to\!\mathbf{G}_1$), $W_2$ ($\Gamma\!\to\!\mathbf{G}_2$),
  $W_3$ ($\Gamma\!\to\!-\mathbf{G}_1\!-\!\mathbf{G}_2$):
  $\|W_i^2+I\|<5\times10^{-2}$,
  $\det(W_i)\in[-1.01,-0.99]$,
  $\|\{W_i,W_j\}\|<5\times10^{-2}$ ($i\neq j$).
\item $\|\{U_i^{\mathrm{proj}},U_j^{\mathrm{proj}}\}\|<10^{-6}$ ($i\neq j$),
  $||\lambda|-1|<10^{-6}$.
\item $\|W_{\mathrm{boundary}}+I\|<2\times10^{-2}$,
  phase sum $\in[6.25,6.32]$ ($=2\pi$ within $0.6\%$, confirming $C=1$).
\item Numerical Computation Code: \texttt{ ... }
\end{enumerate}

\begin{figure}[h]
  \centering
  \includegraphics[width=0.5\textwidth]
    {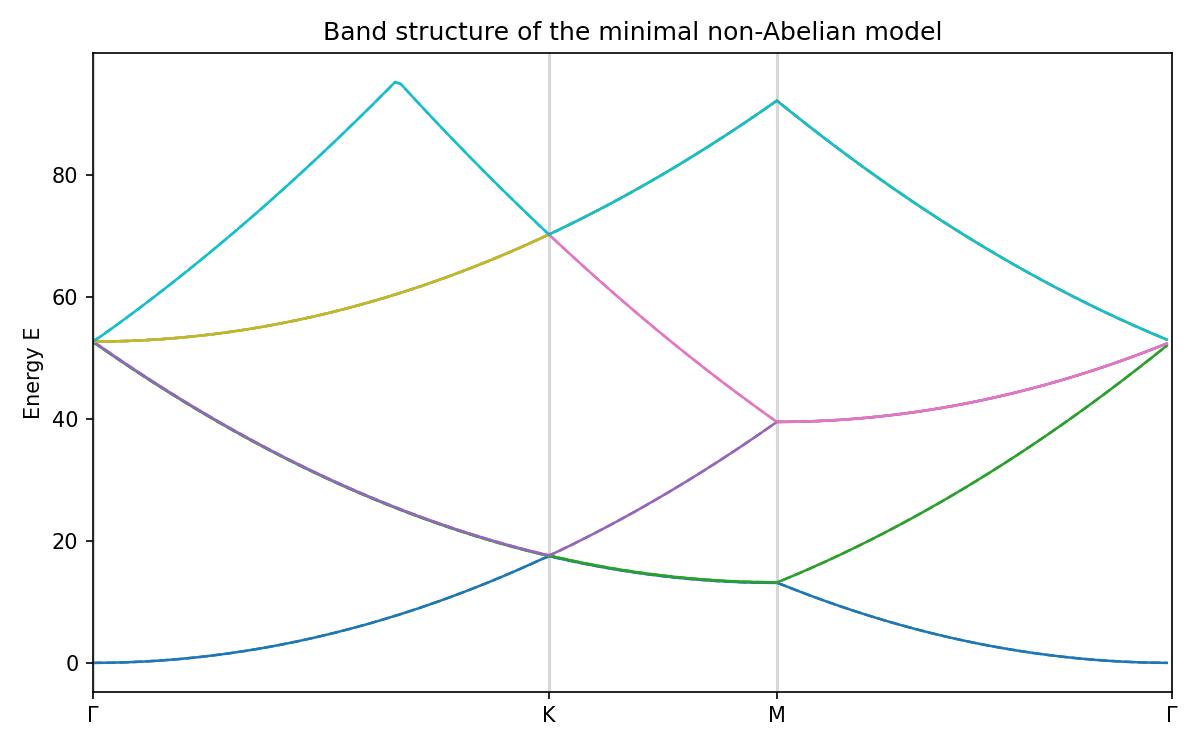}
  \caption{Band structure along $\Gamma$-$K$-$M$-$\Gamma$.
    Solid and dashed: degenerate partner bands; six lowest doublets shown.}
  \label{fig:cmp14_band}
\end{figure}

% ============================================================
\subsection*{Task 3: Spin Texture (Numerical)}

Compute the real-space spin density of the fully occupied lowest doublet,
$S^a(\mathbf{r}) = \frac{1}{N_k^2}\sum_{\mathbf{k}}\sum_{n=1}^2
\langle\psi_{\mathbf{k},n}(\mathbf{r})|\sigma^a|\psi_{\mathbf{k},n}(\mathbf{r})\rangle$.
Using the plane-wave expansion
$|\psi_{\mathbf{k},n}(\mathbf{r})\rangle = \frac{1}{\sqrt{A}}
\sum_{\mathbf{G}} u_{\mathbf{k},n}(\mathbf{G}) e^{i(\mathbf{k}+\mathbf{G})\cdot\mathbf{r}}$,
derive the Fourier form
$S^a(\mathbf{r}) = \sum_{\Delta\mathbf{G}} S^a_{\Delta\mathbf{G}}
e^{i\Delta\mathbf{G}\cdot\mathbf{r}}$,
$S^a_{\Delta\mathbf{G}} = \frac{1}{N_k^2}\sum_{\mathbf{k},n,\mathbf{G}}
u_{\mathbf{k},n}^\dagger(\mathbf{G})\sigma^a u_{\mathbf{k},n}(\mathbf{G}+\Delta\mathbf{G})$.
Implement on $N_k=48$, $N_{\mathrm{cut}}=5$, reconstruct on a $72\times72$
real-space grid ($\mathbf{r}=a_1\mathbf{L}_1+a_2\mathbf{L}_2$, $a_i\in[0,1]$).

\begin{enumerate}
\item Compute $S_x,S_y,S_z$ on the real-space grid.

\item Produce a quiver plot of $\hat{\mathbf{n}}=\mathbf{S}/|\mathbf{S}|$
  (in-plane arrows, colored by $n_z$).
  Fourier-analyze line cuts along $\mathbf{G}_i/|\mathbf{G}_i|$
  to extract the dominant wavelength (expected $\sqrt{3}L/2$).

\item Compute the skyrmion number
  $Q = \frac{1}{4\pi}\int d^2r\;
  \hat{\mathbf{n}}\!\cdot\!(\partial_x\hat{\mathbf{n}}\times\partial_y\hat{\mathbf{n}})$
  via central differences on $(a_1,a_2)$,
  $\partial_x = (L_{2,y}\partial_{a_1}-L_{1,y}\partial_{a_2})/J$,
  $\partial_y = (-L_{2,x}\partial_{a_1}+L_{1,x}\partial_{a_2})/J$,
  $J=|\mathbf{L}_1\times\mathbf{L}_2|=\sqrt{3}L^2/2$.
  Report $Q$ (expected $+2$). Plot the topological density.

\item Connect $Q=+2$ to $C=1$ via the SU(2) gauge structure.
  Show $\langle\mathbf{S}\rangle\approx0$ with $Q\neq0$.
\end{enumerate}

Produce two single-panel figures: (1)~spin direction quiver plot,
(2)~topological density with $Q$ annotated.
Report $Q$, averaged spin, and dominant wavelengths as printed output.

\medskip
\noindent\textbf{Rubrics:}
\begin{enumerate}
\item $\Delta\mathbf{G}$: $|dn_i|\leq10$, 441 components; $S^a(\mathbf{r})
  = \sum_{\Delta\mathbf{G}} S^a_{\Delta\mathbf{G}} e^{i\Delta\mathbf{G}\cdot\mathbf{r}}$
  on $72\times72$ grid. $S^a_{\Delta\mathbf{G}}$ from
  $u^\dagger(\mathbf{G})\sigma^a u(\mathbf{G}+\Delta\mathbf{G})$.
\item $|\langle S_{x,y,z}\rangle|<10^{-4}$; $\max|S_{x,y,z}|\in[0.3,0.9]$.
\item Dominant wavelength $\in[0.83,0.90]$ (expected $\sqrt{3}/2=0.8660$).
\item $Q\in[1.95,2.05]$ for $N_k\geq48$, $N_r\geq72$;
  $|\mathbf{S}(\mathbf{r})|>10^{-6}$ everywhere.
\item Topological density peaks at skyrmion cores;
  $\langle\mathbf{S}\rangle\approx0$ distinct from ferromagnet.
\item $Q=2\leftrightarrow C=1$: SU(2) Berry curvature $\to4\pi$ (real space),
  U(1) trace $\to2\pi$ (momentum space).
\item Numerical Computation Code: \texttt{ ... }
\end{enumerate}

\begin{figure}[h]
  \centering
  \begin{minipage}{0.38\textwidth}
    \centering
    \includegraphics[width=\textwidth]
      {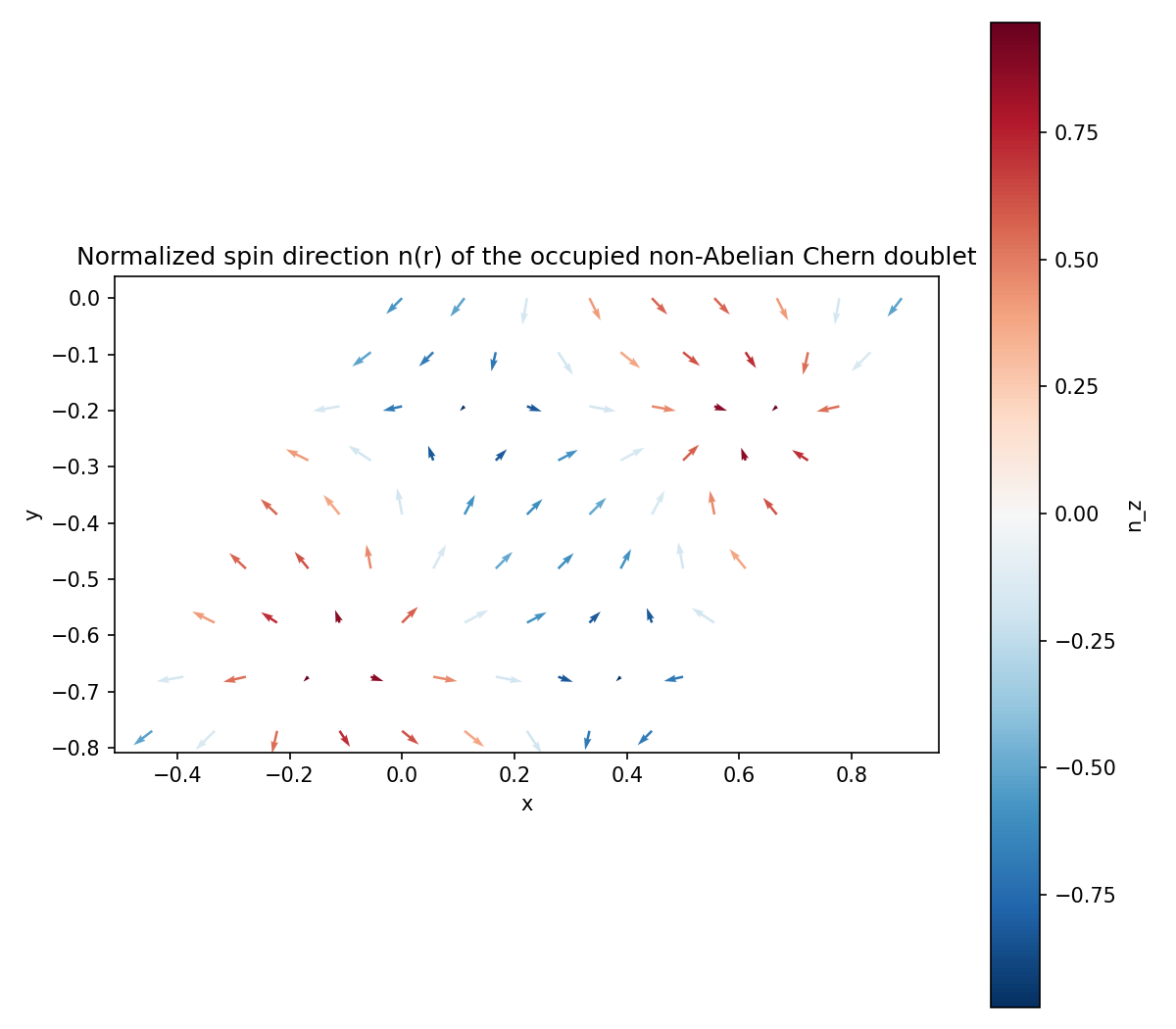}
    \caption{Normalized spin direction
      $\hat{\mathbf{n}}(\mathbf{r})=\mathbf{S}/|\mathbf{S}|$.
      Arrows: in-plane; color: $n_z$.}
    \label{fig:cmp14_spin}
  \end{minipage}
  \hfill
  \begin{minipage}{0.38\textwidth}
    \centering
    \includegraphics[width=\textwidth]
      {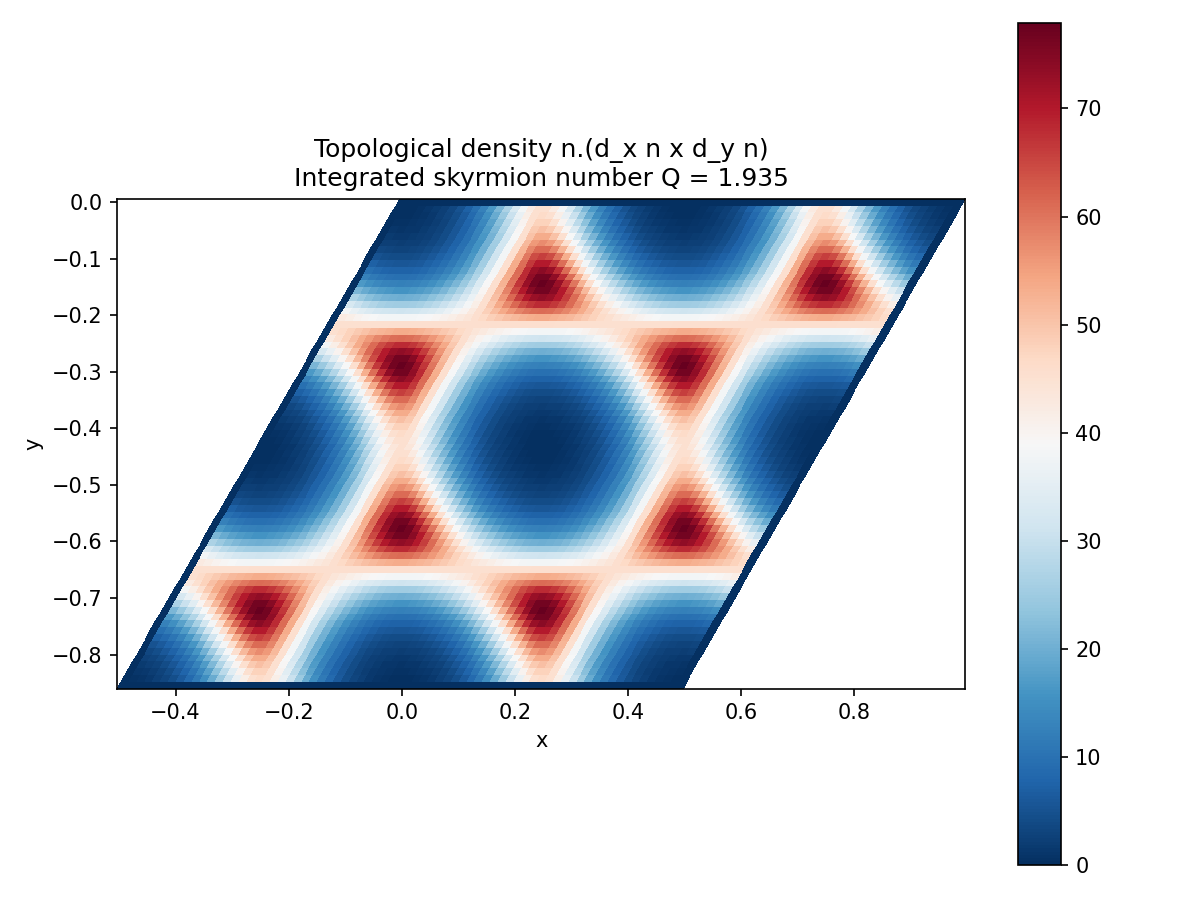}
    \caption{Topological density
      $\hat{\mathbf{n}}\!\cdot\!(\partial_x\hat{\mathbf{n}}\times\partial_y\hat{\mathbf{n}})$
      with integrated $Q$.}
    \label{fig:cmp14_skyrmion}
  \end{minipage}
\end{figure}

\newpage
\section*{Representative Prior Sources}
\label{appendix:textbook}

\textbf{Quantum Field Theory, Renormalization, and High-Energy Theory}

\begin{itemize}
  \item Steven Weinberg, \textit{The Quantum Theory of Fields}, Vol.~1--3
  \item Ashok Das, \textit{Field Theory: A Path Integral Approach}
  \item J.~C. Collins, \textit{Renormalization}
  \item L.~D. Faddeev, \textit{Gauge Fields: An Introduction to Quantum Theory}
  \item Joseph Polchinski, \textit{String Theory}, Vol.~I--II
  \item Ralph Blumenhagen et al., \textit{Introduction to Conformal Field Theory}
  \item Pierre Deligne et al., \textit{Quantum Fields and Strings: A Course for Mathematicians}
\end{itemize}

\textbf{Particle Physics, Standard Model, and Perturbative QCD}

\begin{itemize}
  \item Ta-Pei Cheng and Ling-Fong Li, \textit{Gauge Theory of Elementary Particle Physics}
  \item John C. Collins, \textit{Foundations of Perturbative QCD}
  \item Yuri V. Kovchegov and Eugene Levin, \textit{Quantum Chromodynamics at High Energy}
\end{itemize}

\textbf{Condensed Matter Field Theory and Many-Body Theory}

\begin{itemize}
  \item Alexander Altland and Ben Simons, \textit{Condensed Matter Field Theory}
  \item Piers Coleman, \textit{Introduction to Many-Body Physics}
  \item Henrik Bruus and Karsten Flensberg, \textit{Many-Body Quantum Theory in Condensed Matter Physics}
  \item Nicolas Dupuis, \textit{Field Theory of Condensed Matter and Ultracold Gases}
  \item Subir Sachdev, \textit{Quantum Phases of Matter}
\end{itemize}

\textbf{Statistical Physics and Non-equilibrium Physics}

\begin{itemize}
  \item Mehran Kardar, \textit{Statistical Physics of Particles}
  \item S.~R. de Groot and P. Mazur, \textit{Non-equilibrium Thermodynamics}
\end{itemize}

\textbf{General Relativity, Black Holes, and Cosmology}

\begin{itemize}
  \item Steven Weinberg, \textit{Gravitation and Cosmology}
  \item S. Chandrasekhar, \textit{The Mathematical Theory of Black Holes}
  \item Mike Guidry, \textit{Modern General Relativity: Black Holes, Gravitational Waves, and Cosmology}
  \item Daniel Baumann, \textit{Cosmology}
\end{itemize}

\textbf{Astrophysics, Galactic Dynamics, Accretion, and Compact Objects}

\begin{itemize}
  \item James Binney and Scott Tremaine, \textit{Galactic Dynamics}
  \item Juhan Frank, Andrew King and Derek Raine, \textit{Accretion Power in Astrophysics}
  \item Stuart L. Shapiro and Saul A. Teukolsky, \textit{Black Holes, White Dwarfs and Neutron Stars}
  \item Norman K. Glendenning, \textit{Compact Stars: Nuclear Physics, Particle Physics and General Relativity}
\end{itemize}

%%===========================================================================================%%
%% If you are submitting to one of the Nature Portfolio journals, using the eJP submission   %%
%% system, please include the references within the manuscript file itself. You may do this  %%
%% by copying the reference list from your .bbl file, paste it into the main manuscript .tex %%
%% file, and delete the associated \verb+\bibliography+ commands.                            %%
%%===========================================================================================%%

\bibliography{sn-bibliography}% common bib file
%% if required, the content of .bbl file can be included here once bbl is generated
%%\input sn-article.bbl

\end{document}